\documentclass{article}

\usepackage[final,nonatbib]{neurips_2025}
\usepackage[numbers,sort&compress]{natbib}

\usepackage[utf8]{inputenc}
\usepackage[T1]{fontenc}
\usepackage{hyperref}
\usepackage{url}
\usepackage{booktabs}
\usepackage{amsfonts}
\usepackage{nicefrac}
\usepackage{microtype}
\usepackage{xcolor}
\usepackage{graphicx}
\usepackage{amsmath,amssymb}
\usepackage{subcaption}
\usepackage{tikz} 
\usetikzlibrary{trees}
\usepackage{forest}
\usepackage{nicematrix}   
\usepackage{colortbl}     

\workshoptitle{REO: Advances in Representation Learning for Earth Observation}

\title{HELM: Hierarchical and Explicit Label Modeling with Graph Learning for Multi-Label Image Classification}

\author{%
  Marjan Stoimchev$^{1,2}$ \quad
  Boshko Koloski$^{1,2}$ \\[0.3em]
  Jurica Levatić$^{1}$ \quad
  Dragi Kocev$^{1}$ \quad
  Sašo Džeroski$^{1}$ \\
  $^{1}$ Jo\v{z}ef Stefan Institute, Ljubljana, Slovenia \\
  $^{2}$ Jo\v{z}ef Stefan International Postgraduate School, Ljubljana, Slovenia \\[0.3em]
  \texttt{\{marjan.stoimchev, boshko koloski, jurica.levatic\}@ijs.si} \\
  \texttt{\{dragi.kocev, saso.dzeroski\}@ijs.si}
}

\begin{document}

\maketitle

\begin{abstract}
Hierarchical multi-label classification (HMLC) is essential for modeling complex label dependencies in remote sensing. Existing methods, however, struggle with multi-path hierarchies where instances belong to multiple branches, and they rarely exploit unlabeled data. We introduce HELM (\textit{Hierarchical and Explicit Label Modeling}), a novel framework that overcomes these limitations. HELM: (i) uses hierarchy-specific class tokens within a Vision Transformer to capture nuanced label interactions; (ii) employs graph convolutional networks to explicitly encode the hierarchical structure and generate hierarchy-aware embeddings; and (iii) integrates a self-supervised branch to effectively leverage unlabeled imagery. We perform a comprehensive evaluation on four remote sensing image (RSI) datasets (UCM, AID, DFC-15, MLRSNet). HELM achieves state-of-the-art performance, consistently outperforming strong baselines in both supervised and semi-supervised settings, demonstrating particular strength in low-label scenarios.
\end{abstract}

\let\thefootnote\relax\footnotetext{
\hspace{-1.5em}*\,Correspondence to \texttt{marjan.stoimchev@ijs.si}
}

\section{Introduction}

Hierarchical multi-label classification (HMLC) addresses predictive modeling problems where samples are annotated with multiple labels that are organized in a hierarchy (e.g., a tree or a directed acyclic graph). While incorporating hierarchical information has shown promise in fields like gene function prediction \cite{gene_function_hmlc,gene_function_hmlc_saso} and text categorization \cite{text_hmlc,text_hmlc_v2}, its potential in computer vision, and specifically remote sensing, remains largely unexplored.

\textbf{Limitations.} Modern HMLC methods typically translate the hierarchy into the network design \cite{b-cnn,condition-cnn} or embed hierarchical constraints into the loss function \cite{HRN,c-hmcnn,use_all_labels}. However, they face critical limitations: (i) they often assume single-path hierarchies, failing to model realistic multi-path scenarios where images contain multiple object categories across different branches of the hierarchy; (ii) they underuse the hierarchy; network-based approaches are computationally heavy, and loss-based formulations often miss long-range dependencies; (iii) they focus almost exclusively on supervised learning, ignoring the vast amounts of available unlabeled data. While recent methods have started using Graph Neural Networks (GNNs) \cite{mlc_gcn,hierarchy_gcn_anime}, they remain limited, and semi-supervised learning (SSL) for HMLC in computer vision is practically non-existent.

\textbf{Our Approach.} We propose HELM, a novel semi-supervised framework designed to address these limitations. HELM utilizes a multi-branch architecture with three key components: (i) hierarchy-specific class tokens integrated into a Vision Transformer (ViT) encoder to explicitly model label interactions; (ii) a graph learning branch that uses Graph Convolutional Networks (GCNs) to model dependencies by propagating information through parent-child relationships; (iii) a self-supervised component (BYOL) that leverages unlabeled data to learn robust representations. To our knowledge, HELM is the first semi-supervised HMLC method for images capable of handling complex multi-path hierarchies.

\textbf{Contributions.} Our main contributions are: (1) A novel multi-token transformer architecture that integrates graph-based hierarchical reasoning and self-supervised learning for HMLC. (2) Extensive experiments on four real-world remote sensing datasets demonstrating consistent and significant improvements over baselines and state-of-the-art methods. (3) A framework that effectively leverages unlabeled data, achieving substantial performance gains (up to 37\%) in low-label regimes, which are common in remote sensing applications \cite{safonova2023ten}.

\section{Methodology}

\subsection{Architecture Overview}

\begin{figure}
    \centering
    \begin{minipage}[b]{0.6\textwidth} 
        \centering
        \includegraphics[width=\textwidth]{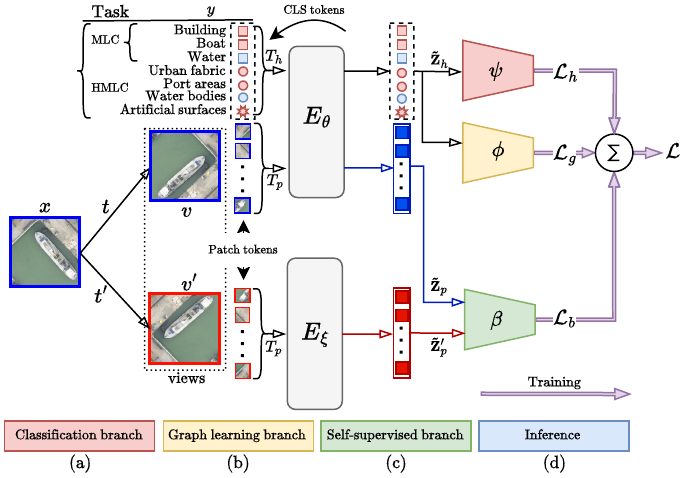}
        \label{fig:method_hmlc}
    \end{minipage}%
    \begin{minipage}[b]{0.37\textwidth} 
        \centering
        \begin{subfigure}[b]{\textwidth}
            \centering
            \includegraphics[width=0.85\textwidth]{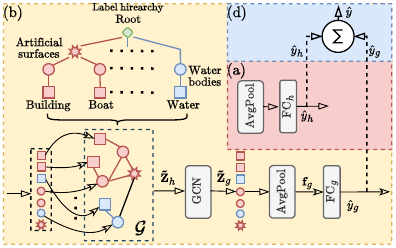}
            \label{fig:gnc_component}
        \end{subfigure}
        \\[1.5em]
        \begin{subfigure}[b]{\textwidth}
            \centering
            \includegraphics[width=0.85\textwidth]{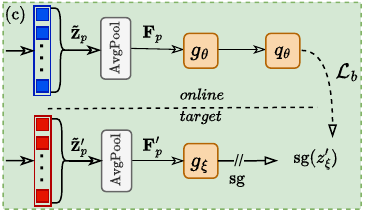}
            \label{fig:byol_component}
        \end{subfigure}
    \end{minipage}
\caption{High-level overview of HELM. The framework integrates a ViT encoder with hierarchy-specific tokens that feed three distinct branches: (a) classification, (b) graph learning with a GCN, and (c) a BYOL self-supervised branch. The losses from each branch are combined to optimize the model end-to-end.}
\label{fig:methodology_overview}
\end{figure}

HELM consists of three branches that are jointly optimized through a composite loss function $\mathcal{L} = \mathcal{L}_s + \mathcal{L}_g + \mathcal{L}_b$ (Figure~\ref{fig:methodology_overview}). These branches are: (i) a classification branch for discriminative learning on labeled data; (ii) a graph learning branch to capture hierarchical dependencies; and (iii) a self-supervised branch to leverage unlabeled data.

\subsection{Encoder with Hierarchy-Specific Tokens}

Given an input image $x \in \mathbb{R}^{C \times W \times H}$, we follow the standard ViT process \cite{vit_original}, dividing it into $N_p$ patches and projecting them to obtain a sequence of patch tokens $T_p \in \mathbb{R}^{N_p \times d}$. To explicitly model the label structure, we introduce $M$ learnable, hierarchy-specific CLS tokens $T_{\texttt{CLS}} \in \mathbb{R}^{M \times d}$, where $M$ is the total number of labels (leaf and intermediate) in the hierarchy. These tokens serve a dual purpose: they provide the output dimensionality for classification and act as initial node embeddings for the graph learning branch.

The tokens are concatenated, $T = [T_{\texttt{CLS}} \| T_p] \in \mathbb{R}^{(M + N_p) \times d}$, and processed through the ViT encoder $E(\cdot,\theta)$. Through the self-attention mechanism, the CLS tokens interact with the patch tokens, allowing them to evolve into semantically meaningful embeddings that represent specific labels.

\subsection{Classification Branch}
This branch performs supervised learning on labeled data. It aggregates the output hierarchy-specific token embeddings $\tilde{\mathbf{z}}_{\texttt{CLS}} \in \mathbb{R}^{M \times d}$ via average pooling to get a unified representation $\mathbf{f}_{\texttt{CLS}}$. This vector is projected by a fully connected layer to the label space. The supervised loss is computed using binary cross-entropy ($\mathcal{H}$) on a batch of $B_l$ labeled samples:
\begin{equation}
\mathcal{L}_s = \frac{1}{B_l} \sum_{i=1}^{B_l} \mathcal{H}(y_i, p_s(y|x_i))
\end{equation}

\subsection{Graph Learning Branch}
To model label dependencies, we construct a directed graph $\mathcal{G} = (\mathcal{V}, \mathcal{E})$ from the label hierarchy. The hierarchy-specific CLS tokens $\tilde{\mathbf{z}}_{\texttt{CLS}}$ serve as initial node features. A GraphSAGE \cite{hamilton2017inductive} operator $\phi(\cdot)$ is applied to propagate information and generate structure-aware embeddings: $\tilde{\mathbf{z}}_g = \phi(\tilde{\mathbf{z}}_{\texttt{CLS}}; \mathcal{G})$. These are then pooled and projected to get predictions. This branch processes the entire batch (labeled and unlabeled), but the loss is computed only on labeled samples, enabling a semi-supervised flow of information through the graph structure:
\begin{equation}
\mathcal{L}_g = \frac{1}{B_l}\sum_{i=1}^{B_l} \mathcal{H}(y_i, p_g(y|x_i))
\end{equation}

\subsection{Self-Supervised Branch}
We integrate Bootstrap Your Own Latent (BYOL) \cite{byol} to leverage unlabeled data. This branch operates on the entire batch in a label-agnostic manner. For each image, two augmented views are created. An online network (sharing the main encoder's weights $\theta$) is trained to predict the representation of the same image from a target network (with weights $\xi$ updated via an exponential moving average of $\theta$). The loss encourages similarity between the predictions of the online network and the projections of the target network:
\begin{equation}
\mathcal{L}_b = 2 - 2 \cdot \frac{\langle q_\theta(g_\theta(\mathbf{F}_p)), g_\xi(\mathbf{F}'_p)\rangle}{\|q_\theta(g_\theta(\mathbf{F}_p))\| \cdot \|g_\xi(\mathbf{F}'_p)\|}
\end{equation}
where $\mathbf{F}_p$ and $\mathbf{F}'_p$ are representations from the two views, $g$ is a projection head, and $q$ is a predictor network.

\section{Results}

We evaluate HELM on four public remote sensing datasets: UCM \cite{ucm_dataset_new}, AID \cite{aid_dataset_new}, DFC-15 \cite{dfc_15}, and MLRSNet \cite{mlrsnet}. These datasets cover a wide range of scene types and hierarchical complexities. Detailed dataset statistics, hierarchy construction, implementation settings, and baseline descriptions are provided in Appendix \ref{app:datasets}–\ref{app:baselines}.

\paragraph{Supervised Results.}
Table~\ref{tab:helm_vs_baselines} shows the supervised results of HELM and its variants. Incorporating hierarchical structure already improves performance, as HMLC surpasses the flat MLC baseline on all datasets. Adding graph reasoning (HELM$_g$) yields further gains, particularly on UCM and DFC-15, confirming the benefit of modeling label dependencies. HELM$_b$ performs best on AID and MLRSNet, indicating that utilizing unlabeled data aids generalization. The complete HELM model achieves the overall best or second-best results, with the highest AU$\overline{\textrm{PRC}}$ on UCM (0.904) and the lowest Ranking Loss across all datasets (0.022, 0.017, 0.006, 0.024).

\begin{table}[h]
\caption{Performance comparison with loss components. Best in \textbf{bold}, second-best \underline{underlined}.}
\centering
\scriptsize
\begin{NiceTabular}{l|ccc|cccc|cccc}
\toprule
& \multicolumn{3}{c|}{\textbf{Loss Components}} & \multicolumn{4}{c|}{\textbf{AU$\overline{\textrm{PRC}}$ ($\uparrow$)}} & \multicolumn{4}{c}{\textbf{Ranking Loss ($\downarrow$)}} \\
\textbf{Method} & \textbf{$\mathcal{L}_s$} & \textbf{$\mathcal{L}_g$} & \textbf{$\mathcal{L}_b$} & \textbf{UCM} & \textbf{AID} & \textbf{DFC-15} & \textbf{MLRSNet} & \textbf{UCM} & \textbf{AID} & \textbf{DFC-15} & \textbf{MLRSNet} \\
\midrule
MLC & \checkmark & & & 0.863 & 0.767 & 0.967 & 0.838 & 0.031 & 0.025 & 0.010 & 0.039 \\
HMLC & \checkmark & & & 0.890 & 0.827 & 0.971 & 0.863 & 0.031 & 0.021 & 0.008 & 0.027 \\
HELM$_g$ & \checkmark & \checkmark & & \underline{0.899} & 0.842 & \textbf{0.979} & 0.869 & \underline{0.024} & \underline{0.019} & \underline{0.007} & \underline{0.025} \\
HELM$_b$ & \checkmark & & \checkmark & 0.885 & \textbf{0.852} & 0.969 & \textbf{0.873} & 0.029 & \underline{0.019} & 0.012 & \underline{0.025} \\
\rowcolor{blue!10}
\textbf{HELM} & \checkmark & \checkmark & \checkmark & \textbf{0.904} & \underline{0.849} & \underline{0.977} & \underline{0.871} & \textbf{0.022} & \textbf{0.017} & \textbf{0.006} & \textbf{0.024} \\
\bottomrule
\end{NiceTabular}
\label{tab:helm_vs_baselines}
\end{table}

\begin{figure}[h]
\centering
\includegraphics[width=0.495\columnwidth]{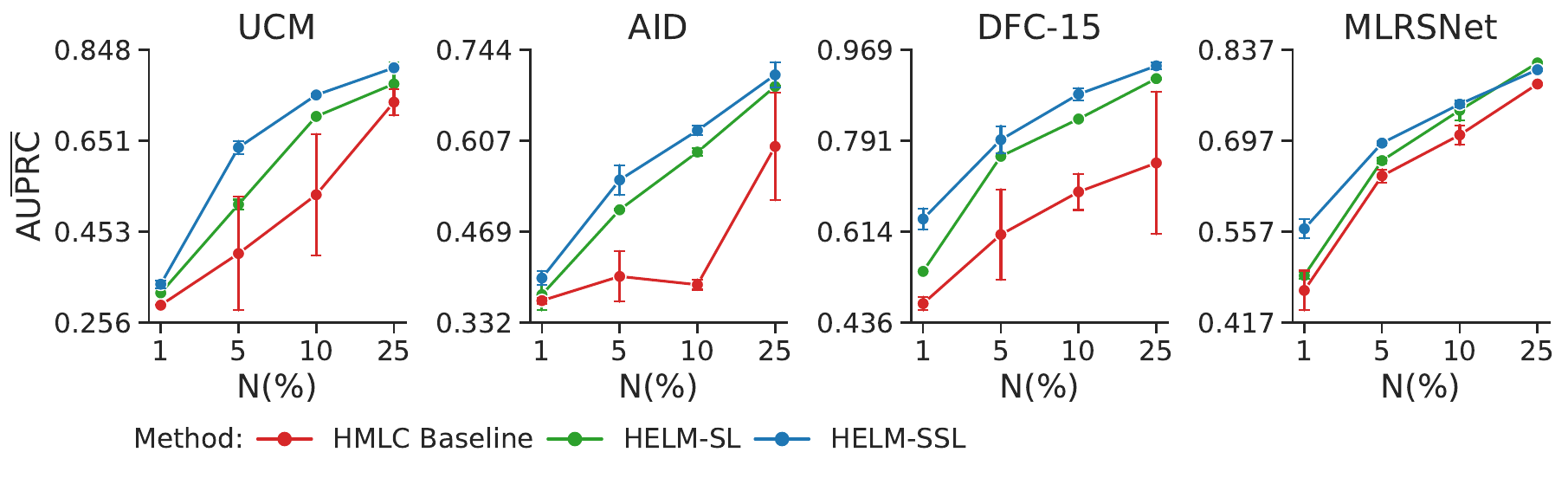}
\hfill
\includegraphics[width=0.495\columnwidth]{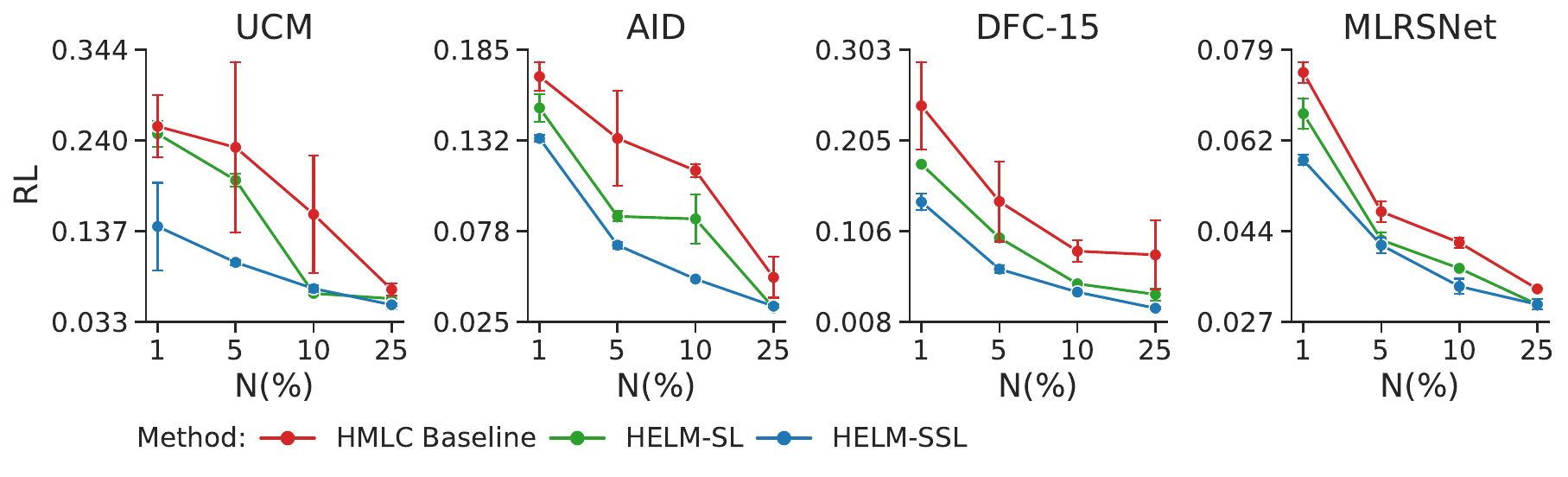}
\caption{Semi-supervised results with different labeled proportions. HELM consistently outperforms the supervised baseline, with the largest gains at 1–5\% labeled data. Full results are available in Appendix~\ref{app:ssl_results}.}
\label{fig:ssl}
\end{figure}

\paragraph{Semi-Supervised Results.}
Figure~\ref{fig:ssl} shows the semi-supervised performance for different amounts of labeled data (1\%, 5\%, 10\%, and 25\%). HELM-SSL consistently surpasses both its supervised variant (HELM-SL) and the supervised HMLC baseline, with the largest AU$\overline{\textrm{PRC}}$ improvements in the lowest label settings. At 1\% supervision, HELM achieves gains of 25.0\% on UCM, 6.6\% on AID, 37.0\% on DFC-15, and 18.5\% on MLRSNet. Complete semi-supervised tables for all datasets and model variants are provided in Appendix~\ref{app:ssl_results}.

\paragraph{Comparison with State of the Art.}
Table~\ref{tab:sota_remote} compares HELM with established hierarchical multi-label classification methods: C-HMCNN~\cite{c-hmcnn}, HiMulConE~\cite{use_all_labels}, and HMI~\cite{hmi}. HELM achieves the highest AU$\overline{\textrm{PRC}}$ and the lowest Ranking Loss on every dataset. It improves upon HiMulConE by 7.2\% on UCM and 10.3\% on AID in AU$\overline{\textrm{PRC}}$, while reducing Ranking Loss by 29.0\% on UCM and 15.0\% on AID. These consistent improvements confirm that hierarchy-specific tokens combined with graph-based reasoning significantly enhance label consistency and predictive accuracy.

\begin{table*}[ht!]
\caption{Comparison with state-of-the-art HMLC methods in the supervised setting. Best in \textbf{bold}, second-best \underline{underlined}.}
\centering
\scriptsize
\setlength{\tabcolsep}{5pt}
\begin{NiceTabular}[color-inside]{lcccccccc}
\toprule
& \multicolumn{4}{c}{\textbf{AU$\overline{\textrm{PRC}}$ ($\uparrow$)}} &
\multicolumn{4}{c}{\textbf{Ranking Loss ($\downarrow$)}} \\
\cmidrule(lr){2-5} \cmidrule(lr){6-9}
\textbf{Method} & \textbf{UCM} & \textbf{AID} & \textbf{DFC-15} & \textbf{MLRSNet} &
\textbf{UCM} & \textbf{AID} & \textbf{DFC-15} & \textbf{MLRSNet} \\
\midrule
C-HMCNN \cite{c-hmcnn} & 0.834 & 0.764 & 0.962 & 0.792 & 0.038 & 0.024 & 0.012 & 0.041 \\
HiMulConE \cite{use_all_labels} & \underline{0.843} & \underline{0.770} & \underline{0.970} & \underline{0.865} &
\underline{0.031} & \underline{0.020} & \textbf{0.006} & \underline{0.035} \\
HMI \cite{hmi} & 0.661 & 0.647 & 0.923 & 0.437 & 0.080 & 0.073 & 0.043 & 0.138 \\
\rowcolor{blue!10}
\textbf{HELM (Ours)} & \textbf{0.904} & \textbf{0.849} & \textbf{0.977} & \textbf{0.871} &
\textbf{0.022} & \textbf{0.017} & \textbf{0.006} & \textbf{0.025} \\
\bottomrule
\end{NiceTabular}
\label{tab:sota_remote}
\end{table*}

\section{Discussion}

We introduced HELM, a novel semi-supervised framework for HMLC that combines hierarchy-specific tokens, graph-based structure encoding, and self-supervised representation learning. The experiments show that HELM consistently improves over strong baselines and achieves state-of-the-art performance on four RSI datasets. By modeling multi-path hierarchies and leveraging unlabeled data, it provides substantial gains, particularly in low-label regimes.

The effectiveness of HELM arises from three complementary design choices. First, hierarchy-specific CLS tokens allow the model to represent each label explicitly and capture fine-grained relationships through self-attention. Second, the GCN propagates information across parent-child links, enriching label embeddings with structural context. Third, the BYOL branch exploits unlabeled imagery to learn generalizable visual features that strengthen the model under limited supervision.

As shown in the Appendix, the graph module adds minimal computational overhead while yielding clear performance benefits. The BYOL branch increases training cost but provides important improvements when labeled data are scarce, which is often the case in remote sensing. 

Future work will focus on automatic hierarchy discovery to remove manual construction, initialization of hierarchy-specific tokens using vision-language models, and extensions to multi-modal inputs such as SAR or multispectral imagery to improve generalization and applicability.

\section*{Acknowledgments}
We acknowledge the financial support of the Slovenian Research and Innovation Agency (ARIS) through the core research programme P2-0103 (Knowledge Technologies), projects J1-3033, J2-2505, J2-4452, J2-4660, J3-3070, J4-3095, J5-4575, J7-4636, J7-4637, and N2-0236. The work of the BK was supported by the Young Researcher grant PR-12394. 

\bibliographystyle{unsrtnat}  
\bibliography{references}     

\appendix
\section{Appendix}

\subsection{Experimental Settings}

\subsubsection{Datasets}\label{app:datasets}

We evaluate HELM across four public remote sensing image datasets: UCM~\cite{ucm_dataset_old,ucm_dataset_new} (2.1k images, 17 leaf labels), AID~\cite{aid_dataset_old,aid_dataset_new} (3k, 17), DFC-15~\cite{dfc_15} (3.3k, 8), and MLRSNet~\cite{mlrsnet} (109k, 60). Table~\ref{tab:dataset_stats_appendix} reports train/test splits, hierarchy sizes, and level-wise cardinality and density.

\begin{table*}[h]
\centering
\caption{Comprehensive statistics for remote sensing datasets. $N$ denotes total images; $N_{train}$ and $N_{test}$ represent training and test set sizes. $|\mathcal{L}|$ indicates the number of unique labels at each hierarchical level (1 to 4), where $\ell$ corresponds to the leaf level and $h$ refers to the total number of labels across all levels. $Card$@Level is the average number of active labels per image at that level (label cardinality), and $Dens$@Level is the mean per-label prevalence at that level (label density).}
\label{tab:dataset_stats_appendix}
\tiny
\setlength{\tabcolsep}{2.5pt}
\resizebox{\linewidth}{!}{%
\begin{tabular}{l l l l| *{6}{c}| *{6}{c}| *{6}{c}}
\toprule
& $N$ & $N_{train}$ & $N_{test}$ & \multicolumn{6}{c|}{$|\mathcal{L}|$} & \multicolumn{6}{c|}{$Card$@Level} & \multicolumn{6}{c}{$Dens$@Level}\\
Dataset & & & & $1$ & $2$ & $3$ & $4$ & ${\ell}$ & ${h}$ & $1$ & $2$ & $3$ & $4$ & ${\ell}$ & ${h}$ & $1$ & $2$ & $3$ & $4$ & ${\ell}$ & ${h}$\\
\midrule
UCM & 2,100 & 1,667 & 433 & 4 & 9 & 17 & - & 17 & 30 & 1.74 & 3.03 & 3.19 & - & 3.19 & 7.96 & 0.44 & 0.34 & 0.19 & - & 0.19 & 0.27 \\
AID & 3,000 & 2,400 & 600 & 4 & 9 & 17 & - & 17 & 30 & 2.18 & 4.51 & 4.93 & - & 4.93 & 11.62 & 0.54 & 0.50 & 0.29 & - & 0.29 & 0.39 \\
DFC-15 & 3,341 & 2,672 & 669 & 3 & 7 & 8 & - & 8 & 17 & 1.61 & 2.73 & 2.80 & - & 2.80 & 6.83 & 0.54 & 0.39 & 0.35 & - & 0.35 & 0.40 \\
MLRSNet & 109,151 & 87,336 & 21,815 & 7 & 15 & 22 & 60 & 60 & 104 & 2.21 & 3.62 & 4.41 & 6.10 & 6.10 & 15.25 & 0.32 & 0.24 & 0.20 & 0.10 & 0.10 & 0.16 \\
\bottomrule
\end{tabular}}
\end{table*}

\noindent\textbf{UCM.} The UC Merced dataset contains 2{,}100 aerial RGB images at $256\times256$ with 17 object-level (leaf) labels~\cite{ucm_dataset_new}. We construct a hierarchical extension by mapping the original categories to CORINE-inspired levels, yielding 30 labels across three levels ($|\mathcal{L}|=\{4,9,17\}$; total $h{=}30$, leaf $\ell{=}17$). This introduces intermediate abstractions while preserving the original leaf space.

\textbf{AID.} The Aerial Image Dataset comprises 3{,}000 images at $600\times600$ with 17 leaf labels~\cite{aid_dataset_new}. We construct a hierarchical version with the same depth and level sizes as UC Merced, resulting in 30 labels across three levels ($|\mathcal{L}|=\{4,9,17\}$; $h{=}30$, $\ell{=}17$). Relative to UC Merced, AID exhibits higher label co-occurrence and per-label prevalence, providing a richer supervisory signal for hierarchical learning.

\textbf{DFC-15.} The 2015 IEEE GRSS Data Fusion Contest dataset provides 3{,}341 image patches at $600\times600$ with 8 leaf categories~\cite{dfc_15}. From the original semantic categories, we construct a compact three-level hierarchy totaling 17 labels ($|\mathcal{L}|=\{3,7,8\}$; $h{=}17$, $\ell{=}8$). Although it contains fewer leaf classes than the other benchmarks, its leaf-level prevalence is highest, yielding dense supervision per image.

\textbf{MLRSNet.} MLRSNet includes 109{,}151 images at $256\times256$ annotated with 60 leaf categories~\cite{mlrsnet}. We construct the deepest hierarchical extension among the four datasets, with 104 labels across four levels ($|\mathcal{L}|=\{7,15,22,60\}$; $h{=}104$, $\ell{=}60$). Images typically carry many active leaf labels, while each individual leaf label is comparatively rare across the corpus, reflecting broad coverage over numerous fine-grained categories and stressing scalability.

\noindent In all cases, the hierarchical versions are constructed from the original multi-label datasets to introduce intermediate semantic levels while retaining the original leaf label sets.

\subsubsection{Hierarchy Construction}\label{app:hierarchy}

\begin{figure}[ht]
\centering
\resizebox{1\textwidth}{!}{%
\begin{forest}
  for tree={
    draw,
    rectangle,
    rounded corners=1pt,
    align=center,
    font=\tiny,
    inner sep=1pt,
    anchor=center,
    l sep=3mm,
    s sep=1mm,
    edge={-},
    parent anchor=south,
    child anchor=north,
    tier/.wrap pgfmath arg={tier #1}{level()},
  }
  [Root
    [Artificial\\Surfaces
      [Urban\\Fabric
        [buildings]
        [mobile-home]
      ]
      [Industrial\\Commercial\\and Transport\\Units
        [airplane]
        [cars]
        [court]
        [dock]
        [ship]
        [storage\\tanks]
      ]
      [Road and Rail\\Networks and\\Associated Land
        [pavement]
      ]
      [Mine Dump\\and Construction\\Sites
        [bare-soil]
      ]
    ]
    [Agricultural\\Areas
      [Arable\\Land
        [field]
      ]
    ]
    [Forest and\\Semi-Natural\\Areas
      [Forests
        [trees]
      ]
      [Shrub and/or\\Herbaceous\\Vegetation\\Associations
        [chaparral]
        [grass]
      ]
    ]
    [Water\\Bodies
      [Inland\\Waters
        [water]
      ]
      [Marine\\Waters
        [sea]
        [sand]
      ]
    ]
  ]
\end{forest}%
}
\caption{Example of constructed label hierarchy for the UCM dataset, derived from the CORINE Land Cover nomenclature. The hierarchy demonstrates the 3-level structure with 4 top-level categories, 9 intermediate-level categories, and 17 leaf-level labels.}
\label{fig:hierarchy_examples_appendix}
\end{figure}
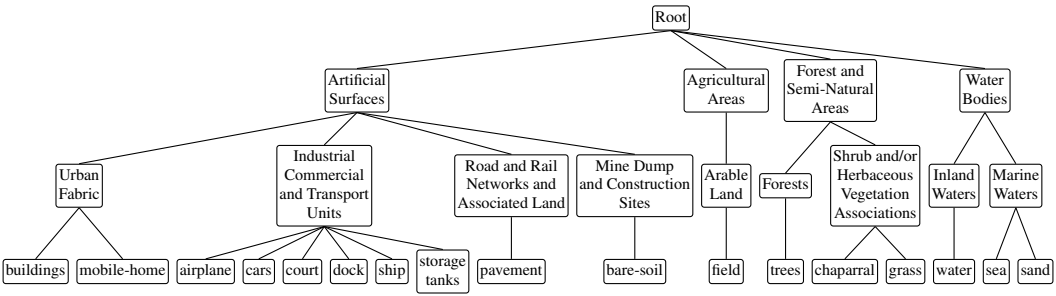

All hierarchical structures were constructed by systematically mapping classes to the CORINE Land Cover nomenclature (CLC)~\cite{corine_landcover}, which provides well-established relationships between land cover types across multiple levels. This enables meaningful hierarchical structures while preserving the multi-label nature at the leaf level. When direct CLC mapping was infeasible, we used ChatGPT-assisted mapping followed by manual validation~\cite{hgclip}. Figure~\ref{fig:hierarchy_examples_appendix} presents a representative example of the constructed hierarchical structure for the UCM dataset.

\subsubsection{Implementation Details}\label{app:implementation}

Our HELM framework is implemented in \texttt{PyTorch Lightning}\footnote{\url{https://github.com/Lightning-AI/pytorch-lightning}}, leveraging the base variant of the Vision Transformer (ViT-B/16) with a patch size of $16 \times 16$ pixels. The ViT backbone is initialized with pre-trained ImageNet weights to provide strong visual priors. For the graph learning component, we employ \texttt{PyTorch Geometric (PyG)}\footnote{\url{https://pytorch-geometric.readthedocs.io/en/latest/}} with a two-layer GraphSAGE network, enabling efficient message passing within the label hierarchy. The self-supervised BYOL component is integrated using \texttt{LightlySSL}\footnote{\url{https://github.com/lightly-ai/lightly}}, a specialized library for self-supervised learning in computer vision.

Models are trained on RGB images resized to $224 \times 224$ pixels using the unified learning objective. The hierarchy-specific CLS tokens produce latent representations of dimension $d = 768$ after average pooling. For the supervised components (classification and graph branches), we apply weak augmentations, including horizontal and vertical flips and Gaussian blur. For the BYOL component, we employ both weak and strong augmentations, with the latter incorporating color jittering, affine transformations, random cropping, and random erasing to strengthen representation learning.

Training is conducted for 100 epochs using the AdamW optimizer with a base learning rate of $1\times10^{-4}$, cosine annealing scheduler, and a mini-batch size of 16. Mixed-precision training is utilized to accelerate computation and reduce memory footprint. All experiments are performed on four NVIDIA A100 GPUs (40~GB each).

To ensure full reproducibility, all hierarchical label structures constructed for each dataset will be released as YAML configuration files, providing the exact parent–child mappings used in the graph module. These YAML files, together with the training scripts and configuration examples, will be included in the public code release accompanying the final version of this paper.

\subsubsection{Evaluation Strategy}\label{app:evaluation}

In the semi-supervised setting, we vary the labeled ratio in \{1, 5, 10, 25\}\%, using the remainder as unlabeled. For each configuration, we perform three independent runs with different random seeds to ensure statistical reliability. Performance is evaluated on a fixed test set using exclusively leaf labels, regardless of the training configuration. This ensures fair comparison across all methods, including those that use hierarchical information during training (HMLC, HELM variants) and those that do not (MLC baseline). We report AU$\overline{\text{PRC}}$ and ranking loss, averaged over the three independent runs.

\subsubsection{Computational Resources}

All experiments were conducted on four NVIDIA A100 GPUs equipped with 40 GB memory each.

\subsubsection{Evaluation Metrics}

We employ two performance metrics to evaluate the effectiveness of the methods: the average area under the precision-recall curve (AUPRC) and the ranking loss measure. Given that the tasks involve multi-label classification and hierarchical multi-label classification, we use a variant of AUPRC--the area under the micro-averaged Precision-Recall curve (AU$\overline{\textrm{PRC}}$). These metrics are chosen because they are independent of classification thresholds and provide a reliable assessment of performance.

In this context, $\overline{\textrm{Prec}}$ represents the proportion of predicted labels that are correct, while $\overline{\textrm{Rec}}$ corresponds to the proportion of actual labels in the dataset that are correctly predicted. These values are calculated as follows:

\begin{equation}
\overline{\textrm{Prec}} = \frac{\sum_i \textrm{TP}_i}{\sum_i \textrm{TP}_i + \sum_i \textrm{FP}_i},
\end{equation}

\begin{equation}
\overline{\textrm{Rec}} = \frac{\sum_i \textrm{TP}_i}{\sum_i \textrm{TP}_i + \sum_i \textrm{FN}_i},
\end{equation}
where $i$ iterates over all classes. By varying the decision threshold, an average Precision-Recall (PR) curve is generated.

Ranking loss (RL), measuring the average fraction of incorrectly ordered label pairs for a given example, is defined as

\begin{equation}
\text{RL} = \frac{1}{N} \sum_{i=1}^N \frac{1}{|Y_i^+| \cdot |Y_i^-|} \sum_{y_p \in Y_i^+} \sum_{y_n \in Y_i^-} \mathbb{I}(f_i(y_p) \leq f_i(y_n)),
\end{equation}
where \(N\) is the number of examples, \(Y_i^+\) and \(Y_i^-\) are the sets of positive and negative labels for the \(i\)-th example, \(f_i(y)\) is the predicted score for label \(y\), and \(\mathbb{I}(\cdot)\) is the indicator function, which returns 1 if the condition is true and 0 otherwise. The loss is weighted by the sizes of \(Y_i^+\) and \(Y_i^-\), penalizing each pair of misordered labels equally. A lower ranking loss indicates better performance, as fewer label pairs are incorrectly ranked.

To compute these metrics, we use the readily available implementations from \texttt{scikit-learn} \cite{scikit}.

\subsubsection{Compared Methods}\label{app:baselines}

\textbf{HELM Variants and Baseline Methods.} To systematically evaluate the contribution of each component, we analyze different variants of HELM by selectively including loss terms from the overall objective. Table~\ref{tab:helm_variants} summarizes all evaluated configurations and corresponding loss components.

\begin{table}[h]
\centering
\caption{HELM variants and baseline methods with their loss component configurations for systematic evaluation.}
\label{tab:helm_variants}
\begin{tabular}{llcccl}
\toprule
\textbf{Method Variant} & \textbf{Notation} & $\mathcal{L}_s$ & $\mathcal{L}_g$ & $\mathcal{L}_b$ & \textbf{Task} \\
\midrule
MLC Baseline & MLC & \checkmark & & & \{SL\} \\
HMLC Baseline & HMLC & \checkmark & & & \{SL\} \\
HELM (Graph-only) & HELM$_g$ & \checkmark & \checkmark & & \{SL, SSL\} \\
HELM (BYOL-only) & HELM$_b$ & \checkmark & & \checkmark & \{SL, SSL\} \\
HELM (Full) & HELM & \checkmark & \checkmark & \checkmark & \{SL, SSL\} \\
\bottomrule
\end{tabular}
\end{table}

We establish two baseline methods to isolate the impact of hierarchical information: the MLC baseline, which uses only leaf labels during training and performs standard multi-label classification, and the HMLC baseline, which incorporates the complete hierarchical label set during training but treats all labels independently without exploiting structural relationships (essentially a flat approach to hierarchical classification). Both baselines utilize only the supervised classification loss $\mathcal{L}_s$, omitting the graph and self-supervised components.

The HELM variants systematically introduce additional components to assess their individual and combined contributions. HELM$_g$ adds graph-based dependency modeling to capture hierarchical label relationships, while HELM$_b$ incorporates self-supervised learning through BYOL to leverage unlabeled data. The complete HELM model integrates all three components, providing a unified framework that combines supervised classification, hierarchical reasoning, and self-supervised representation learning.

This experimental design enables precise measurement of how each component contributes to performance, particularly in scenarios with limited labeled supervision. All methods are evaluated using exclusively leaf labels of testing examples to ensure fair comparison, regardless of training configuration.

\textbf{State-of-the-Art Methods.} We also compare HELM against three state-of-the-art HMLC methods: C-HMCNN \cite{c-hmcnn}, HiMulConE \cite{use_all_labels}, and HMI \cite{hmi}. These methods represent a diverse range of approaches to hierarchical multi-label classification, highlighting the strengths and limitations of current techniques.

C-HMCNN \cite{c-hmcnn} ensures consistency between parent and child predictions through a modified binary cross-entropy (BCE) loss, providing flexibility in handling complex hierarchies. Although originally designed for tabular data, we adapted it for image-based tasks by replacing its tabular encoder with our vision transformer encoder. HiMulConE \cite{use_all_labels} adopts a two-stage approach: it first employs a hierarchy-preserving contrastive loss to learn label-aware embeddings and then trains a classifier on leaf nodes using cross-entropy and softmax. Since the original method utilizes a cross-entropy loss designed for single-label classification at the leaf level, we replaced it with a BCE loss to support scenarios where multiple labels may exist at the leaf level. HMI \cite{hmi} utilizes a hyperbolic Poincaré ball model to encode logical relationships, such as implication and exclusion, through geometric constraints. This approach achieves high consistency with fewer dimensions, enhancing computational efficiency. However, similar to C-HMCNN, HMI was originally designed for tabular data and we adapted it for image-based tasks by incorporating our vision transformer encoder to enable a direct comparison.

For a fair evaluation, all methods are implemented using our vision encoder, and we utilize the official code repositories of each method, setting hyperparameters according to the original authors' recommendations.

\subsection{Complete Semi-Supervised Learning Results}\label{app:ssl_results}

Table~\ref{tab:ssl_complete_appendix} presents comprehensive semi-supervised learning results across all labeled data proportions for each dataset. The results demonstrate HELM's effectiveness in leveraging unlabeled data, with particularly strong performance gains in low-label scenarios.

\begin{table*}[ht!]
\tiny
\centering
\setlength{\tabcolsep}{1.8pt}
\caption{Evaluation of HELM's individual loss components in the semi-supervised learning setting compared to the supervised HMLC baseline. Models are trained with varying proportions of labeled data and evaluated on a fixed test set. Results are reported using AU$\overline{\textrm{PRC}}$ (higher is better) and Ranking Loss (lower is better), averaged over three runs. The best performance for each dataset and labeled data proportion is highlighted in bold.}
\label{tab:ssl_complete_appendix}

\begin{NiceTabular}{l 
cccc@{\hskip 8pt}
cccc@{\hskip 8pt}
cccc@{\hskip 8pt}
cccc@{\hskip 8pt}
cccc}

\toprule

AU$\overline{\textrm{PRC}}(\uparrow)$ & 
\multicolumn{4}{c@{\hskip 8pt}}{UCM} & 
\multicolumn{4}{c@{\hskip 8pt}}{AID} & 
\multicolumn{4}{c@{\hskip 8pt}}{DFC-15} & 
\multicolumn{4}{c@{\hskip 8pt}}{MLRSNet} &
\multicolumn{4}{c}{Avg. ranks} \\

&
\multicolumn{4}{c@{\hskip 6pt}}{N (\%)} & 
\multicolumn{4}{c@{\hskip 6pt}}{N (\%)} & 
\multicolumn{4}{c@{\hskip 6pt}}{N (\%)} & 
\multicolumn{4}{c@{\hskip 6pt}}{N (\%)} &
\multicolumn{4}{c}{N (\%)} \\

Methods & 
1 & 5 & 10 & 25 & 
1 & 5 & 10 & 25 & 
1 & 5 & 10 & 25 & 
1 & 5 & 10 & 25 &
1 & 5 & 10 & 25 \\

\cmidrule(l{1pt}r{8pt}){2-5}
\cmidrule(l{1pt}r{8pt}){6-9}
\cmidrule(l{1pt}r{8pt}){10-13}
\cmidrule(l{1pt}r{8pt}){14-17}
\cmidrule(l{1pt}r{2pt}){18-21}

HMLC Baseline & 
0.268 & 0.357 & 0.479 & 0.691 & 
0.366 & 0.410 & 0.382 & 0.564 & 
0.481 & 0.619 & 0.692 & 0.766 & 
0.470 & 0.643 & 0.716 & 0.787 &
3.3 & 3.8 & 3.5 & 3.8 \\

HELM$_b$\cellcolor{white} & 
0.290 & 0.554 & 0.656 & 0.770 & 
0.362 & 0.461 & 0.557 & 0.671 & 
0.522 & 0.717 & 0.824 & 0.909 & 
0.380 & 0.643 & 0.709 & 0.788 &
3.3 & 3.0 & 3.0 & 2.8 \\

HELM$_g$\cellcolor{white} & 
0.269 & 0.532 & 0.714 & 0.799 & 
0.375 & 0.485 & 0.597 & 0.680 & 
0.619 & 0.792 & 0.875 & 0.916 & 
0.422 & 0.657 & 0.707 & 0.781 &
2.5 & 2.0 & 2.5 & 2.5 \\

\rowcolor{blue!11}
HELM\cellcolor{white} & 
\textbf{0.335} & \textbf{0.651} & \textbf{0.730} & \textbf{0.807} & 
\textbf{0.390} & \textbf{0.533} & \textbf{0.631} & \textbf{0.701} & 
\textbf{0.659} & \textbf{0.790} & \textbf{0.885} & \textbf{0.934} & 
\textbf{0.557} & \textbf{0.666} & \textbf{0.743} & \textbf{0.816} &
\textbf{1.0} & \textbf{1.3} & \textbf{1.0} & \textbf{1.0} \\

\midrule

RL$(\downarrow)$ & 
\multicolumn{4}{c@{\hskip 8pt}}{UCM} & 
\multicolumn{4}{c@{\hskip 8pt}}{AID} & 
\multicolumn{4}{c@{\hskip 8pt}}{DFC-15} & 
\multicolumn{4}{c@{\hskip 8pt}}{MLRSNet} &
\multicolumn{4}{c}{Avg. ranks} \\

Methods & 
1 & 5 & 10 & 25 & 
1 & 5 & 10 & 25 & 
1 & 5 & 10 & 25 & 
1 & 5 & 10 & 25 &
1 & 5 & 10 & 25 \\

\cmidrule(l{1pt}r{8pt}){2-5}
\cmidrule(l{1pt}r{8pt}){6-9}
\cmidrule(l{1pt}r{8pt}){10-13}
\cmidrule(l{1pt}r{8pt}){14-17}
\cmidrule(l{1pt}r{2pt}){18-21}

HMLC Baseline &
0.266 & 0.232 & 0.178 & 0.080 & 
0.169 & 0.122 & 0.137 & 0.062 & 
0.232 & 0.164 & 0.101 & 0.088 & 
0.074 & 0.048 & 0.042 & 0.033 &
3.0 & 3.8 & 3.8 & 3.5 \\

HELM$_b$\cellcolor{white} & 
0.247 & 0.125 & 0.086 & 0.057 & 
0.216 & 0.088 & 0.059 & 0.041 & 
0.179 & 0.092 & 0.058 & 0.033 & 
0.098 & 0.051 & 0.043 & 0.034 &
2.3 & 2.5 & 2.5 & 2.8 \\

HELM$_g$\cellcolor{white} & 
0.307 & 0.127 & 0.068 & 0.050 & 
0.155 & 0.076 & 0.056 & 0.039 & 
0.137 & 0.067 & 0.039 & 0.031 & 
0.084 & 0.043 & 0.041 & 0.033 &
3.8 & 2.3 & 2.3 & 2.0 \\

\rowcolor{blue!11}
HELM\cellcolor{white} & 
\textbf{0.137} & \textbf{0.106} & \textbf{0.072} & \textbf{0.051} & 
\textbf{0.133} & \textbf{0.071} & \textbf{0.048} & \textbf{0.033} & 
\textbf{0.132} & \textbf{0.065} & \textbf{0.037} & \textbf{0.023} & 
\textbf{0.061} & \textbf{0.041} & \textbf{0.036} & \textbf{0.030} &
\textbf{1.0} & \textbf{1.3} & \textbf{1.3} & \textbf{1.3} \\

\bottomrule

\end{NiceTabular}
\end{table*}

The complete results demonstrate that HELM consistently outperforms both the HMLC baseline and ablated variants across all datasets and labeled data proportions. The performance gains are particularly pronounced in extremely low-resource scenarios at 1-5\% labeled data. At 1\% labeled data, HELM achieves substantial AU$\overline{\text{PRC}}$ improvements of 25.0\% on UCM, 6.6\% on AID, 37.0\% on DFC-15, and 18.5\% on MLRSNet compared to the HMLC baseline. As the proportion of labeled data increases, the performance gap narrows but remains consistently favorable for HELM across both metrics, demonstrating that the self-supervised component continues to provide benefits even with more abundant supervision.

\subsubsection{Computational Efficiency Analysis}\label{app:run_time}

\begin{figure}[t]
    \centering
    \includegraphics[width=0.6\linewidth]{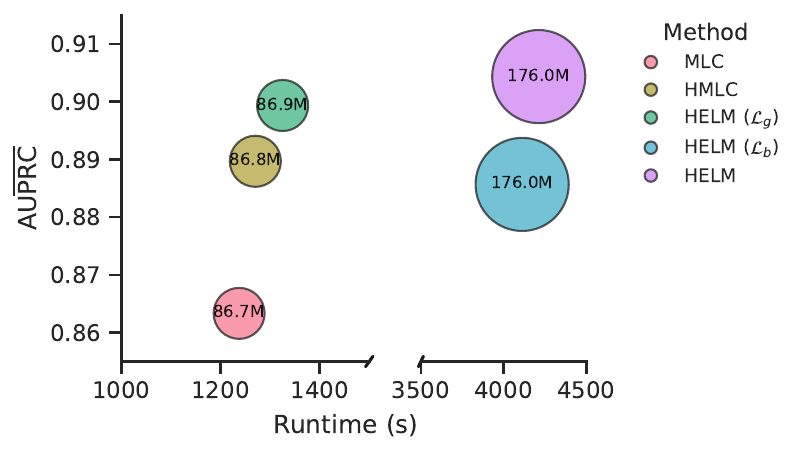}
    \caption{Comparison of training times, performance, and parameter counts for the baseline methods and HELM on the UCM dataset. The size of each bubble corresponds to the number of parameters (in millions).}
    \label{fig:training_time}
\end{figure}

Figure~\ref{fig:training_time} illustrates the computational trade-offs inherent in our framework design, evaluated on the representative UCM dataset. 
The graph learning component proves remarkably efficient, adding only 107K parameters while delivering substantial performance gains. In contrast, the BYOL component introduces significant computational overhead due to its dual-encoder architecture.

\subsection{Qualitative Evaluation}

This section presents a qualitative evaluation of HELM compared to state-of-the-art methods using the UCM dataset.

\subsubsection{Visualization of Learned Embeddings}

The embeddings are derived from the hierarchy-specific CLS tokens produced by the final layer of the encoder. Since our encoder serves as a common backbone architecture across all methods, the comparison between learned embeddings is both fair and direct, ensuring that differences arise solely from the design and learning mechanisms of the methods. We use Uniform Manifold Approximation and Projection (UMAP) to visualize the high-dimensional embeddings.

To quantify embedding quality, we use Normalized Mutual Information (NMI), which evaluates the alignment between clusters formed by embeddings and ground-truth labels. Before computing NMI, we apply $k$-Nearest Neighbors clustering on the original embeddings, with the number of clusters set to match the number of labels at each hierarchical level. We report the average NMI calculated by averaging NMI values across different levels of the hierarchy. This approach provides a comprehensive evaluation of both fine-grained (leaf-level) and broader (ancestor-level) clustering quality.

\begin{figure*}
    \centering
    \includegraphics[width=1\textwidth]{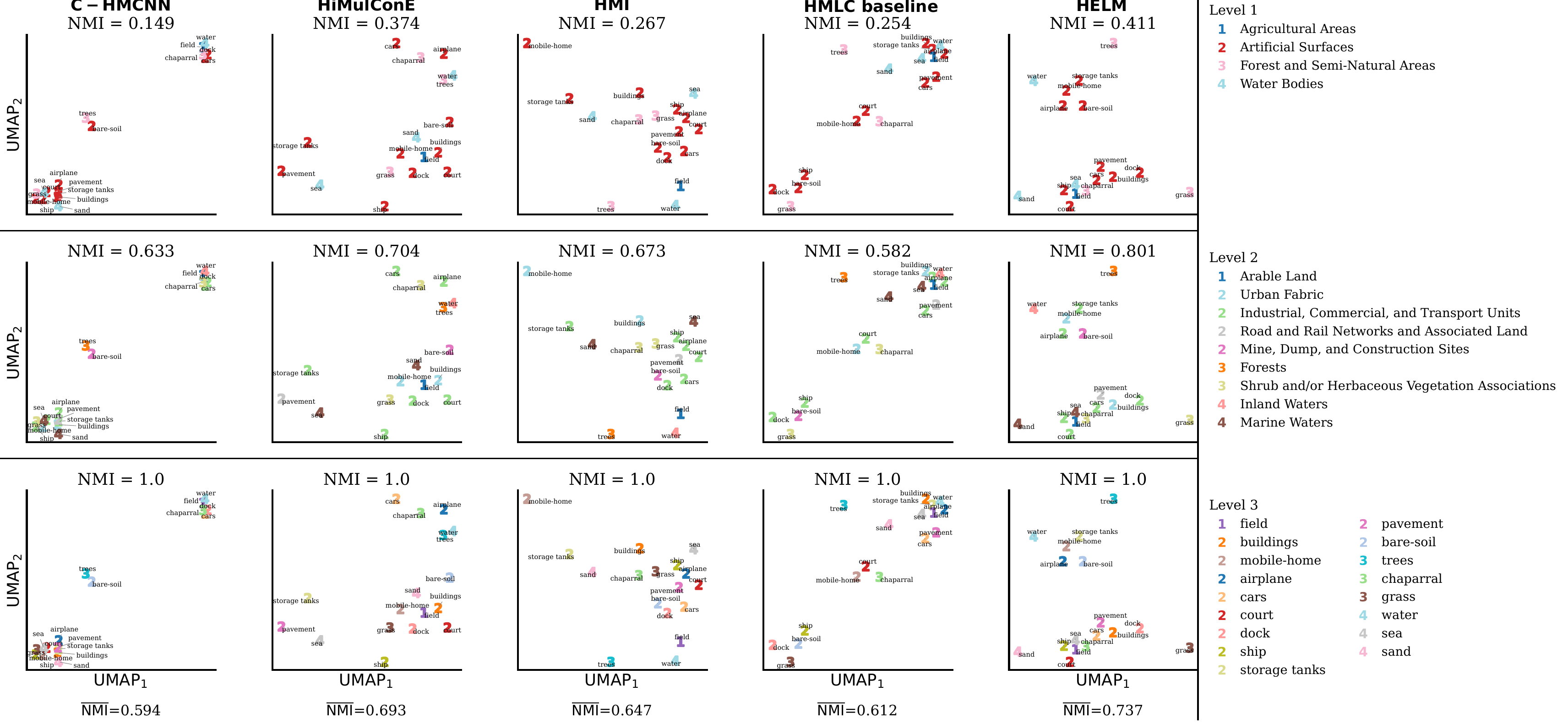}
    \caption{Comparison of 2-D UMAP embeddings between HELM and state-of-the-art methods for the UCM dataset. The learned embeddings are colored based on different levels of the UCM label hierarchy. The visualization is based on embeddings corresponding to leaf labels, while the color coding reflects the grouping and relationships at each hierarchical level. The NMI values are reported for each method, where higher values indicate better alignment between clusters and ground truth labels, reflecting the quality of hierarchical embeddings.}
    \label{fig:umap}
\end{figure*}

Figure~\ref{fig:umap} presents the UMAP analysis results across the hierarchical levels of the UCM dataset. HELM with hierarchical modeling delivers the best results, achieving NMI values of 0.411 at the first level and 0.801 at the second level of the hierarchy, with an average NMI of $\overline{\mathrm{NMI}} = 0.737$ across all levels. The embeddings produced are well-structured and form distinct clusters that align with the hierarchical label relationships, demonstrating the effectiveness of explicitly modeling hierarchical dependencies. This approach allows the model to capture both coarse-grained and fine-grained relationships, resulting in superior clustering quality and predictive performance. In contrast, the HMLC baseline achieves $\overline{\mathrm{NMI}}$ of 0.612, placing it as the second-lowest performing method in the comparison. This highlights the significant limitations of excluding hierarchical relationships, which restrict the model's ability to organize embeddings meaningfully. While HiMulConE delivers competitive results, it falls short of capturing the hierarchical structure as effectively as HELM with hierarchical modeling.

\subsection{Limitations}

While HELM outperforms state-of-the-art models, our approach has limitations. The BYOL branch is computationally intensive, significantly increasing runtime and parameter count due to its dual-encoder architecture. This computational cost is justified by the enhanced representation quality and improved generalization capabilities observed across our experimental evaluation, particularly in low-label scenarios where annotation costs are prohibitive.

As shown in the UMAP analysis (Figure~\ref{fig:umap}), while HELM outperforms state-of-the-art methods and generates well-defined clusters that align with the hierarchical structure of labels, it occasionally fails to consistently cluster hierarchical levels. For instance, labels such as "sand" and "water", which share the same parent label "Water Bodies" at higher levels of the hierarchy, are well-clustered individually but fail to fully reflect their shared broader grouping. This suggests that although HELM excels at modeling fine-grained relationships, there is room for improvement in capturing and representing broader hierarchical structures more consistently.


\newpage
\section*{EurIPS Paper Checklist}

\begin{enumerate}

\item {\bf Claims}
    \item[] Question: Do the main claims made in the abstract and introduction accurately reflect the paper's contributions and scope?
    \item[] Answer: \answerYes{}
    \item[] Justification: The abstract and Introduction clearly describe the three main components of HELM: hierarchy-specific tokens, graph-based reasoning, and self-supervised learning. They also state the expected improvements in both supervised and semi-supervised settings. These claims are directly supported by the experimental results shown in Table~\ref{tab:helm_vs_baselines} and Table~\ref{tab:sota_remote}, as well as by the learning curves in Figure~\ref{fig:ssl}, which confirm the accuracy and scope of the stated contributions.

\item {\bf Limitations}
    \item[] Question: Does the paper discuss the limitations of the work performed by the authors?
    \item[] Answer: \answerYes{}
    \item[] Justification: The paper explicitly discusses both computational and methodological limitations. Appendix~\ref{app:run_time} presents a quantitative analysis of computational efficiency, highlighting the added cost of the BYOL branch and the minimal overhead of the graph module. The dedicated \textit{Limitations} subsection further reflects on representational shortcomings in capturing high-level hierarchical groupings and the increased runtime associated with the dual-encoder design (see Fig.~\ref{fig:umap}). These discussions provide a balanced view of HELM’s trade-offs in performance and scalability.

\item {\bf Theory assumptions and proofs}
    \item[] Question: For each theoretical result, does the paper provide the full set of assumptions and a complete (and correct) proof?
    \item[] Answer: \answerNA{}
    \item[] Justification: The paper is empirical and methodological without formal theorems or proofs.

\item {\bf Experimental result reproducibility}
    \item[] Question: Does the paper fully disclose all the information needed to reproduce the main experimental results of the paper to the extent that it affects the main claims and/or conclusions of the paper (regardless of whether the code and data are provided or not)?
    \item[] Answer: \answerYes{}
    \item[] Justification: All experimental details required for reproduction are provided. The datasets, hierarchy construction process, model configurations, training setup, metrics, and evaluation protocols are described in Appendix~\ref{app:datasets}–\ref{app:baselines}, \ref{app:implementation}, and \ref{app:evaluation}, with model variants summarized in Table~\ref{tab:helm_variants}. To ensure full reproducibility, the exact hierarchical label structures used for each dataset will be released as YAML configuration files alongside the code.

\item {\bf Open access to data and code}
    \item[] Question: Does the paper provide open access to the data and code, with sufficient instructions to faithfully reproduce the main experimental results, as described in supplemental material?
    \item[] Answer: \answerNo{}
    \item[] Justification: Public datasets are cited and described (Appendix~\ref{app:datasets}), but an anonymized code release and scripts are not included in the current submission. We plan to release code and detailed run scripts upon acceptance.

\item {\bf Experimental setting/details}
    \item[] Question: Does the paper specify all the training and test details (e.g., data splits, hyperparameters, how they were chosen, type of optimizer, etc.) necessary to understand the results?
    \item[] Answer: \answerYes{}
    \item[] Justification: Implementation and training details, augmentations, optimizer, schedules, batch sizes, epochs, and evaluation setup are provided in Appendix~\ref{app:implementation} and \ref{app:evaluation}.

\item {\bf Experiment statistical significance}
    \item[] Question: Does the paper report error bars suitably and correctly defined or other appropriate information about the statistical significance of the experiments?
    \item[] Answer: \answerYes{}
    \item[] Justification: We report averages and standard deviations over three independent runs with different random seeds in the learning curves (Fig.~\ref{fig:ssl}), where the variability bands make the stability of the methods clear. All tables report mean values only, while per-run results and standard deviations are omitted for brevity but are reflected in the learning curves.

\item {\bf Experiments compute resources}
    \item[] Question: For each experiment, does the paper provide sufficient information on the computer resources (type of compute workers, memory, time of execution) needed to reproduce the experiments?
    \item[] Answer: \answerYes{}
    \item[] Justification: We provide details on the computational setup (on single A100 40GB GPU) and report a dedicated computational efficiency analysis in Appendix~\ref{app:run_time} and Figure~\ref{fig:training_time}. 
    The analysis quantifies runtime, parameter counts, and trade-offs across model variants on the UCM dataset. 
    For simplicity, runtime profiling is presented on a single dataset, but the relative scaling trends remain consistent across all benchmarks. 
    Training times are comparable between runs, with the graph branch adding negligible overhead and the BYOL component introducing a predictable increase in runtime due to its dual-encoder structure.

\item {\bf Code of ethics}
    \item[] Question: Does the research conducted in the paper conform, in every respect, with the NeurIPS Code of Ethics \url{https://neurips.cc/public/EthicsGuidelines}?
    \item[] Answer: \answerYes{}
    \item[] Justification: The work uses public remote sensing datasets with citations, no sensitive personal data, and follows standard evaluation practices. No violations of the Code of Ethics were identified.

\item {\bf Broader impacts}
    \item[] Question: Does the paper discuss both potential positive societal impacts and negative societal impacts of the work performed?
    \item[] Answer: \answerNo{}
    \item[] Justification: While we do not include a dedicated broader impacts section, the paper discusses potential positive implications in environmental monitoring, urban planning, and climate assessment within the Introduction and Discussion. Given the foundational and non-sensitive nature of this research, no direct negative societal impacts are anticipated.

\item {\bf Safeguards}
    \item[] Question: Does the paper describe safeguards that have been put in place for responsible release of data or models that have a high risk for misuse (e.g., pretrained language models, image generators, or scraped datasets)?
    \item[] Answer: \answerNA{}
    \item[] Justification: The work does not release high-risk foundation models or scraped web datasets. It evaluates task models on public benchmarks.

\item {\bf Licenses for existing assets}
    \item[] Question: Are the creators or original owners of assets (e.g., code, data, models), used in the paper, properly credited and are the license and terms of use explicitly mentioned and properly respected?
    \item[] Answer: \answerNo{}
    \item[] Justification: All datasets and prior methods used in this work are properly cited in the main text and appendix. However, the specific dataset and code licenses are not explicitly listed in the current version for brevity. We will include the corresponding license information for each asset in the camera-ready version to ensure full compliance and transparency.

\item {\bf New assets}
    \item[] Question: Are new assets introduced in the paper well documented and is the documentation provided alongside the assets?
    \item[] Answer: \answerNA{}
    \item[] Justification: The paper does not introduce new datasets or public model checkpoints at the time of submission. Upon acceptance, all configuration files, hierarchical mappings, and trained model weights will be documented and released to support full reproducibility.

\item {\bf Crowdsourcing and research with human subjects}
    \item[] Question: For crowdsourcing experiments and research with human subjects, does the paper include the full text of instructions given to participants and screenshots, if applicable, as well as details about compensation (if any)? 
    \item[] Answer: \answerNA{}
    \item[] Justification: The research does not involve crowdsourcing or human subjects.

\item {\bf Institutional review board (IRB) approvals or equivalent for research with human subjects}
    \item[] Question: Does the paper describe potential risks incurred by study participants, whether such risks were disclosed to the subjects, and whether Institutional Review Board (IRB) approvals (or an equivalent approval/review based on the requirements of your country or institution) were obtained?
    \item[] Answer: \answerNA{}
    \item[] Justification: The research does not involve human subjects and therefore does not require IRB review.

\item {\bf Declaration of LLM usage}
    \item[] Question: Does the paper describe the usage of LLMs if it is an important, original, or non-standard component of the core methods in this research?
    \item[] Answer: \answerYes{}
    \item[] Justification: We used ChatGPT in two limited and fully disclosed ways: (i) to assist in hierarchy mapping for classes without direct CORINE Land Cover alignment (Appendix~\ref{app:hierarchy}), and (ii) for light paraphrasing and linguistic refinement of manuscript text to improve clarity and readability. The model was not involved in designing methods, conducting experiments, or interpreting results.

\end{enumerate}

\end{document}